\documentclass[review]{elsarticle}

\usepackage{hyperref}
\usepackage{float}
\usepackage{verbatim} %comments
\usepackage{apalike}
\restylefloat{figure}
\restylefloat{table}
\usepackage{xcolor}

\usepackage{amsmath,amssymb,amsfonts}
\usepackage{algorithmic}
\usepackage{graphicx}
\usepackage{siunitx}
\usepackage{subcaption}
\usepackage{layouts}
\usepackage{booktabs}

\newcommand{\R}{\mathbb{R}}
\renewcommand{\vec}[1]{\mathbf{#1}}
\let\oldhat\hat
\renewcommand{\hat}[1]{\oldhat{\mathbf{#1}}}

\journal{Expert Systems with Applications}

\biboptions{authoryear}

\begin{document}
\begin{frontmatter}

\begin{titlepage}
\begin{center}
\vspace*{1cm}

\textbf{ \large Formation Control for UAVs Using a Flux Guided Approach}

\vspace{1.5cm}

% Author names and affiliations
John Hartley$^{a,d}$ (john.hartley@ed.ac.uk), Hubert P. H. Shum$^{a,b}$ (hubert.shum@durham.ac.uk), Edmond S. L. Ho$^{a}$ (e.ho@northumbria.ac.uk), He Wang$^{c}$ (h.e.wang@leeds.ac.uk), Subramanian Ramamoorthy$^{d}$ (s.ramamoorthy@ed.ac.uk) \\

\hspace{10pt}

\begin{flushleft}
\small  
$^a$ Department of Computer and Information Sciences, Northumbria University, Newcastle upon Tyne, UK \\
$^b$ Department of Computer Science, Durham University, Durham, UK \\
$^c$ School of Computing, University of Leeds, Leeds, UK \\
$^d$ School of Informatics, University of Edinburgh, Edinburgh, UK

\vspace{1cm}
\textbf{Corresponding Author:} \\
Hubert P. H. Shum \\
Department of Computer Science, Durham University, Durham, UK \\
Tel: +44 1913342724 \\
Email: hubert.shum@durham.ac.uk

\end{flushleft}        
\end{center}
\end{titlepage}

\title{Formation Control for UAVs Using a Flux Guided Approach}

\author[label1,label4]{John Hartley}
\ead{john.hartley@ed.ac.uk}

\author[label1,label2]{Hubert P. H. Shum \corref{cor1}}
\ead{hubert.shum@durham.ac.uk}

\author[label1]{Edmond S. L. Ho}
\ead{e.ho@northumbria.ac.uk}

\author[label3]{He Wang}
\ead{h.e.wang@leeds.ac.uk}

\author[label4]{Subramanian Ramamoorthy}
\ead{s.ramamoorthy@ed.ac.uk}

\cortext[cor1]{Corresponding author.}
\address[label1]{Department of Computer and Information Sciences, Northumbria University, Newcastle upon Tyne, UK}
\address[label2]{Department of Computer Science, Durham University, Durham, UK}
\address[label3]{School of Computing, University of Leeds, Leeds, UK}
\address[label4]{School of Informatics, University of Edinburgh, Edinburgh, UK}

\begin{abstract}
%\printinunitsof{in}\prntlen{\textwidth}
Existing studies on formation control for unmanned aerial vehicles (UAV) have not considered encircling targets where an optimum coverage of the target is required at all times. Such coverage plays a critical role in many real-world applications such as tracking hostile UAVs. This paper proposes a new path planning approach called the Flux Guided (FG) method, which generates collision-free trajectories for multiple UAVs while maximising the coverage of target(s). Our method enables UAVs to track directly toward a target whilst maintaining maximum coverage. Furthermore, multiple scattered targets can be tracked by scaling the formation during flight. FG is highly scalable since it only requires communication between sub-set of UAVs on the open boundary of the formation's surface. 
\textcolor{black}{%To do this we reformulate an existing least-squares flux minimisation problem as a constrained optimisation problem with feasible solutions defined by the geometry of the UAV formation. We solve this problem using a generic sequential quadratic programming algorithm.
Experimental results further validate that FG generates UAV trajectories $1.5 \times$ shorter than previous work and that trajectory planning for 9 leader/follower UAVs to surround a target in two different scenarios only requires 0.52 seconds and 0.88 seconds, respectively.} The resulting trajectories are suitable for robotic controls after time-optimal parameterisation; we demonstrate this using a 3d dynamic particle system that tracks the desired trajectories using a PID controller.

\end{abstract}


\end{frontmatter}

\section{Introduction}
\label{sec:introduction}

% the problem domain and why its important
The formation control of unmanned aerial vehicles (UAV) has been of interest to researchers for several decades. A major reason is that in many applications complex tasks can be accomplished more efficiently by multiple UAVs  \cite{uavswarm}. For example, multiple UAVs can improve autonomous navigation by working together to share power and multi-sensor hardware, even though the payload of the individual drones is limited \cite{lu2018}. The ability to plan feasible trajectories that have non-colliding properties is crucial to operating multiple UAVs.

% The research problem we are trying to solve
Formation control is a particularly important topic for tracking and surrounding remote targets using multiple UAVs. Realistic applications include tracking and neutralisation of aerial threats \cite{proliferateddrones}, as well as search and rescue operations \cite{7353783}. Successful operation in these scenarios requires the formation to maximally face toward the target such that the target can be observed and surrounded more easily. Also, the geometry of the path taken by the UAVs should occupy a minimum volume such that the UAVs can operate within restrictive environments and move efficiently. 

% Why existing work cannot solve it, with a few key references
Existing work introduces simple control systems to control multiple UAVs for tracking targets \cite{uavswarm}. However, there is little coordination among the UAVs in the formation, and many stand-by UAVs have no practical effect. The risk of hostile UAV attacks can be mitigated by small-scale formations. For example, a hostile drone can be tackled by a few drones carrying a dragnet \cite{7991391}. However, these systems do not consider the dynamics of the hostile drones and have a limited coverage of their target. Smart formation controls informed by the dynamic environment and targets are introduced \cite{PAUL20081453}, but they are typically pre-defined algorithms focusing on multi-UAV co-ordinations. More recent methods for UAV formation control utilise leader-follower frameworks for more effective controls \cite{Zhang2018}, and decentralised frameworks for better robustness \cite{Ma2018}. However, few studies have investigated tracking and surrounding multiple targets whilst maintaining full coverage of the target during flight.

% What we propose, our insights, how they solve the problem, method overview
In this paper, we propose a UAV formation control scheme to track and surround dynamic targets. Our work is inspired by the properties of static electric fields, and in particular electric flux. Flux is a smooth potential field suitable for UAV control systems, and computationally inexpensive to compute over simple geometries \cite{boundary}. This is extremely important since a UAV formation spans a region of space rather than a point as in single robot control. Minimisation of electric flux has been used to control the motion of 3d objects in computer graphics \citet{Wang2013} and single robots \citet{Ivan2015}. We adapt these flux methodologies for UAV formation control by minimising the flux through the formation generated by one or more targets. We propose a novel constrained minimisation of the flux through the formation. This enables the UAVs to encircle their target with maximum coverage during flight. This is highly desirable for mitigating hostile UAV threats since the threats can be optimally observed at all times. We show that FG derives more efficient paths toward the target than the baseline approach \citet{Wang2013}, reduces the complexity of the approach given by \cite{Ivan2015}, and eliminates the free parameters in the soft constraints of both methods.

% Experiment overview
We validate our approach by conducting experiments by simulating multiple UAV flights with a proportional–integral–derivative (PID) controller. We demonstrate that the trajectories planned using the flux-guided controller are feasible for robotic PID controls. \textcolor{black}{We support our claims with a wide variety of scenarios including one or more targets, and with multiple conditions of the starting rotation and position of the formation. A detailed sensitivity test is conducted to showcase the robustness of our system.} We further show that UAV paths created by our method are superior to that of \cite{Wang2013}. Finally, we show how path planning can be implemented in a leader-follower framework to minimize the computational cost. We have also included a supplementary video that visualises the UAV motion in 3d.

% Contributions
The contributions of the work are two-fold: 
\begin{itemize}
\item We introduce a new method called the Flux Guided (FG) approach for UAV formation control by adapting \cite{Wang2013, Ivan2015}. We propose a constrained minimisation method to solve for geometry paths that encircle a target with maximum coverage during the flight. The resultant system outperforms \cite{Wang2013} and has a lower complexity than \cite{Ivan2015}.
\item
We show that these paths are suitable for robotic control of UAV quadcopters, after time-optimal path parameterisation (TOPP) with kinematic constraints, using a 3d dynamic particle simulation and a PID controller.
\end{itemize}

The rest of the paper is arranged as follows. In Section \ref{sec:related}, we review work related to this research. In Section \ref{sec:flux}, we explain how we adapt flux optimisation for UAV formation control. In Section \ref{sec:constrain}, we highlight our novel constrained minimisation component that enables robust real-world UAV tracking and surrounding. In Section \ref{sec:experiment}, we validate the effectiveness of the proposed flux-guided system with different experiments. Finally, in Section \ref{sec:conclusions}, we conclude this work and discuss future research directions.

\section{Related Work}
\label{sec:related}

\subsection{Artificial Potential Fields}
Several studies have used artificial potential fields (APF) to guide single agents through known environments \citep{Anipulators1985}. Typically, attractive and repulsive potential fields are constructed to represent the target and obstacles in the environment. The direction of travel of an agent is determined by calculating the force on the agent by gradient descent of the sum of the attractive and repulsive potentials \citep{PAUL20081453, Wang2008a, 6858777, Chen2015}. 

Local minimas in the potential field can be problematic since there is no global solution to the path planning problem. Previous studies have resolved the local minima issues 
%by carefully constructing potential fields, such that a global minimum at the target exists. Alternatively, they can be resolved
by escaping local minima with global optimisation methods such as simulated annealing \citep{JanabiSharifi1993, Engineering2001, Zhu2006}. The main limitation of these methods is that they often require ad-hoc solutions, which may not be robust to dynamic and complex environments like scenarios with multiple moving targets.

In contrast, electric flux is always smooth and does not have local minima. \cite{Wang2013} used the idea of minimising electric flux to develop controllers for computer graphics, and showed that the wrapping of cloth around an object could be effectively modelled. In this work, we draw inspiration from \cite{Wang2013} and adapt the flux minimisation methodology for UAV controls. We show that the paths generated by our method are smooth and compatible with PID controllers.

For formation control of UAVs, the risk of collisions must be minimised and the relative distance between any two UAVs should be maintained. In \cite{barnett16coordinated}, the topology of the environment is considered such that agents can be distributed into possible pathways while following the potential field to minimize congestion. Spatial constraints such as those introduced by \cite{henry14interactive} allow agents to keep a reasonable distance while maintaining the overall formation. Such approaches can be extended to an arbitrary number of agents through control signal abstraction \citep{shen18datadriven}. This research proposes a constrained minimisation method to maintain the distances between UAVs.

% jh
\subsection{Leader-follower Approaches}
APF methods have also been used to plan paths for multiple agents moving in formation. A popular approach is to develop a leader-follower relationship among the agents in the formation \citep{Kowdiki2012,Zhang2018,Li:7904763}. The leader's motion is determined by the APF approach and the followers' motions are guided by the leader using a tracking or control mechanism. The positions of the followers can be determined by maintaining a separation distance and angle with respect to the motion of the leader \citep{Kowdiki2012}, or by using additional potential fields for the leaders and followers \citep{Zhang2018}. Robust adaptive controllers have also been proposed to estimate the indeterminate relative distance between the leader and the followers \citep{Li:7904763}. These approaches are commonly used due to the simplicity of the implementation. However, they can be vulnerable as the leader UAV is a single point of failure. This can be mitigated using virtual leaders, \citep{Leonard2001,PAUL20081453}, and implicit leaders \citep{HE2018327}.

The main drawback of the single leader approach is that the leader's direction of movement defines the orientation of the formation \citep{LU201967}. When the formation should face the target, this behaviour is problematic since a change in the direction of the leader causes a large change in the structure of the formation, which happens when the formation adapts to the new direction of travel. Additional information such as visual images \citep{Chen2015} is needed to adjust the relative orientation between the leader and the followers.% to tackle this problem.
Furthermore, the risk of collision is increased without a feedback mechanism between the leaders and followers. This assignment problem can be solved in a distributed manner \cite{Morgan2016}. However, in such a case, the formation is forced to break down during the re-assignment stage and therefore does optimally cover the target during the flight.

This research simplifies the formation control problem by specifying the location of the target and the formation, which includes both the trajectories and orientations (i.e. facing the target) for all of the UAVs. Also, flux-based methods ensure that the paths are collision-free. 

\subsection{Decentralised Approaches}
Decentralised algorithms for dynamic encirclement of moving targets are presented in \cite{Marasco2013, Franchi2016, Ma2018}. These works show that a simulated group of UAVs can encircle a moving target, and that the tracking can be improved with a model of the target's future states. The disadvantage of these methods is that the UAVs track toward optimal final states and therefore the coverage of the formation is not optimal during flight.

\cite{Wang2013} presents an alternative approach where a target is covered by an open surface defined by a set of agents. The paths of the agents are solved by maximising the electric flux through the open surface defined by the topology of the agents. The agents are located in an electric flux which is generated by a static charge at target's location \cite{Mainprice}. This approach can be extended to extract a common representation to facilitate comparison between 3d objects with different shapes and topological \cite{Sandilands:VRST2014}. Collision free paths are always obtained because colliding agents reduce the area of the open boundary and thus reduce the total flux. Maintaining optimal coverage during flight limits the control over the formation's scale. This leads to optimal coverage but at a very long range from the target. Therefore, this method is not suitable when the target should be closely surrounded. Also, the paths generated by the method have been seen to be indirect and therefore inefficient for locating targets.

\cite{Ivan2015} proposes a centralised scheme to encircle static objects with deformable robotic surfaces. A target flux is specified and successive iterations are used to minimise the error between the current flux and the target flux to generate the encirclement trajectories \citep{exotica}. A limitation of this work is that the minimisation relies on free parameters, which have to be tuned for each encirclement context. Also, the approach is not scalable since the Jacobian of the flux must be computed for every vertex on the deformable surface.

While a decentralised implementation of our flux-guided formation controller is out of the scope of this paper, the concept of flux is compatible with decentralised approaches. We discuss such potential future extensions at the end of this paper.

\section{Flux Optimisation for Formation Trajectory Planning}
\label{sec:flux}
\cite{Wang2013} proposes the flux method for computer graphics.  A cloak modelled using a collection of points acting under spring constraints is controlled to cover a target. An iterative approach updates the positions of the points. It requires that a positive change in the electric flux occurs through the open surface of the cloak due to an electric charge at the target's position. Collision free paths are obtained since a collapse of the surface reduces the overall flux. The method is superior to a rigid body planner with six degrees of freedom since the objective function is designed to maximise coverage at every point of approach. The cloak can cover arbitrary geometrically complex objects since the superposition of the electric fields due to a collection of targets is always harmonic.

In this work, we extend \cite{Wang2013} to the path planning problem of covering a target with a formation of UAVs. The system can be used for a wide range of scenarios. For example, maximum coverage and direct line sight of targets during flight make it an ideal choice for countering hostile UAVs. Other possible use cases include protecting static assets such as airports and buildings. The system can protect moving objects such as vessels in a maritime environment when the planner is updated iteratively.

In this section, we provide an overview of the method in \cite{Wang2013}, reappropriating it for the UAV formation planning problem. We refer to the original formulation as least-squares flux (LS). Firstly, we show the calculation for the flux in Section \ref{sec:FluxEva} through the surface of the formation. Secondly, the update method is presented in Section \ref{sec:updating}. Finally, we illustrate the novel property that FG does not require the coordinates of every UAV in the formation in Section \ref{sec:ReduceDim}.

\subsection{Flux Evaluation} \label{sec:FluxEva}
The first step in LS is to consider a formation and the surface it defines such that the flux can be calculated. Here, we adapt the methodology to direct a formation of UAVs toward a target. 

\begin{figure}[!htb]
    \begin{center}
        \includegraphics[width=0.5\linewidth]{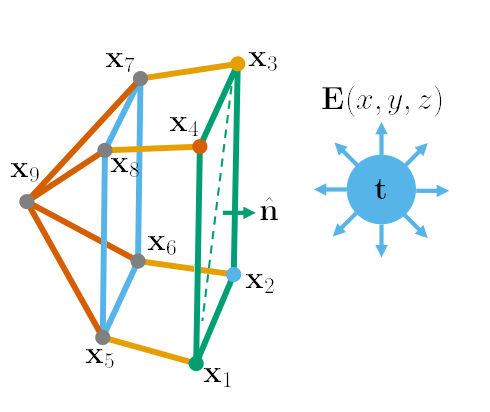}
        \caption{{Points on a open hemispherical surface representing a hemispherical formation. The target is a point charge at position $\vec{t}$ that generates an electric field $\vec{E}(x, y, z)$. A flat surface $S_1$ with a normal vector $\hat{n}$ is bounded by UAVs $\chi_1$, ..., $\chi_4$ at points $\vec{x}_1$, ..., $\vec{x}_4$, and a curved $S_2$ surface is bounded by UAVs $\chi_1$, ..., $\chi_5$, ..., $\chi_9$ at points $\vec{x}_1$, ..., $\vec{x}_5$, ..., $\vec{x}_9$.}}
    \label{fig:hemisphere}
    \end{center}
\end{figure}

We demonstrate LS and FG using a hemispherical formation in this work. Although the approach extends to any set of positions or number of UAVs that lie on an open surface. Figure \ref{fig:hemisphere} shows the hemispherical formation used in this work. Here, we give an example of a formation consisting of nine UAVs placed on a hemispherical surface of radius $r$. The formation can be extended to any number of UAVs. The UAVs are denoted $\chi_1$, ..., $\chi_9$, and are located at positions $\vec{x}_1$, ..., $\vec{x}_9$ respectively.

The UAVs are divided into two groups. The first lie on the forward-facing surface that inscribes a circle of radius $r$, while the second group lie on the curved surface of formation behind the forward-facing surface. In our example, $\chi_1, \chi_2, \chi_3, \chi_4$ form a square. The surface of the square is denoted $S_1$ and has a normal vector $\hat{n}$. The UAVs $\chi_5, \chi_6, \chi_7, \chi_8$ form a square which inscribes a circle translated $\frac{r}{2}$ in the negative $\hat{n}$ direction. The UAV $\chi_9$ is $r$ from the first circle in the negative $\hat{n}$ direction. We define the closed surface of the formation, $S = S_1 + S_2$. Where $S_2$ is the curved surface, which passes through the points $\vec{x}_1, \vec{x}_2, \vec{x}_3, \vec{x}_4, \vec{x}_5, \vec{x}_6, \vec{x}_7, \vec{x}_8, \vec{x}_9$.

The electric flux, $\Phi$, through the hemispherical surface of the formation is given by:
\begin{equation}
    \Phi = \iint_{S_2} \vec{E} \cdot \hat{n}   \,dA \
    \label{equation:gauss-law}
\end{equation}
where $\vec{E}$ is the electric field generated by the target, $\hat{n}$ is a normal vector to the surface $S_2$, and $dA$ is an infinitesimal area on $S_2$.

The exact flux through $S_2$ can be efficiently evaluated by triangulating the surface at the UAVs vertices and summing through resultant flux through each triangle. The flux or solid angle through each triangle due to an isotropic source is given by Equation \ref{eq:flux-flux-triangle} \cite{4121581}:
\begin{align}
    \tan{\frac{\Phi}{2}} = \frac{(\vec{v_1} \times \vec{v_2})\cdot \vec{v_3}}
    { |\vec{v_1}||\vec{v_2}||\vec{v_3}| + 
    (\vec{v_1}\cdot\vec{v_2})|\vec{v_3}| +
    (\vec{v_1}\cdot\vec{v_3})|\vec{v_2}| +
    (\vec{v_2}\cdot\vec{v_3})|\vec{v_1}|
    } 
    \label{eq:flux-flux-triangle}
\end{align}
where $\vec{v_1} = \vec{t} - \vec{p_1}$, $\vec{v_2} = \vec{t} - \vec{p_2}$, $\vec{v_3} = \vec{t} - \vec{p_3}$, $\vec{t}$ is the location of the source, and $\vec{p_1}$, $\vec{p_2}$, $\vec{p_3}$ are the position vectors of the vertices of triangle.

\subsection{UAV Positions Updating} \label{sec:updating}
The geometric path of each UAV is found iteratively. Since we have a function of the flux in terms of the positions of the UAVs, we define a total differential that approximates the local change in flux with position:
\begin{equation}
    \Delta \Phi = \vec{J}_{\phi}\vec{\Delta x} ,
    \label{equation:delta-flux}
\end{equation}
where $\vec{\Delta x} = (\vec{\Delta x}_1 \cdot \hat{i}, \vec{\Delta x}_2 \cdot \hat{i},..., \vec{\Delta x}_4 \cdot \hat{i},..., \vec{\Delta x}_1 \cdot \hat{j}, ..., \vec{\Delta x}_4 \cdot \hat{k}$) denotes a vector of the change in position of each UAV,  $\vec{J}_\phi$ is the Jacobian of the flux with respect to the changes in the positions of the UAVs, and $\hat{i}$, $\hat{j}$, $\hat{k}$ are the Cartesian unit vectors.

Now, we define a Tikhonov regularized least-squares problem that minimises the error in the total differential and a target change in flux, $\phi_r$ with a regularisation term, $\alpha$ is a soft constraint that increases the influence of the flux term:
\begin{align}
    L(\vec{\Delta x}) = \vec{\Delta x}^T\vec{\Delta x} + \alpha(\phi_r - \vec{J}_\phi\vec{\Delta x})^2 
    \label{equation:ls-cost}
\end{align}
where $L(\vec{\Delta x})$ is the quadratic cost function in $\vec{\Delta x}$.

\subsection{Reducing the Dimensionality of the Planning Problem} \label{sec:ReduceDim}

The reduction in dimensionality of the planning (Equation \ref{equation:ls-cost}) can be shown by considering Gauss's law for electric fields:
\begin{equation}
    \iint_{S} \vec{E} \cdot \hat{n} \,dA\ = \frac{Q}{\epsilon_0} 
\end{equation}
where $\vec{E}$ is the electric field,  $dA$ is an infinitesimal area on the Gaussian surface, $\hat{n}$ is the unit normal to the surface, $Q$ is the charge, and $\epsilon_0$ is the free-space dielectric permittivity.

The law states that the flux through a closed surface is proportional to the enclosed charge, $Q$ and the flux. However, in this problem, the flux through the closed surface is zero, since the charged target is outside of the surface as shown in Figure \ref{fig:hemisphere}. Splitting the integral across the surface $S_1$, Equation \ref{equation:split-integral}, and $S_2$, we find the magnitudes of flux through $S_1$ and $S_2$.

\begin{equation}
    \iint_{S_1} \vec{E} \cdot \hat{n} \,dA\ + \iint_{S_2} \vec{E} \cdot \hat{n} \,dA\ = 0 
    \label{equation:split-integral}
\end{equation}
\begin{equation}
    |\Phi_1| = |\Phi_2|
    \label{equation:flux-equality}
\end{equation}

From Equation \ref{equation:flux-equality}, the magnitudes of the flux through $S_1$ and $S_2$ are equal. Therefore, the dimensionality of the planning (Equation \ref{equation:ls-cost}) is reduced by evaluating the flux through the UAVs on the boundary of $S_1$ only. For example, in Figure \ref{fig:hemisphere}, there are four UAVs on $S_1$, and nine UAVs on the $S_2$ boundary. Considering only the boundary UAVs, the dimensionality of the $\vec{\Delta{x}}$ space in Equation \ref{equation:ls-cost} is reduced from 27 to 12. The reduction has the advantage that many UAVs can be placed on $S_2$ boundary without increasing the dimensionality of Equation \ref{equation:ls-cost} whilst retaining the property of maximum coverage of the target during flight.

{A minimum of three UAVs are required to compute the $S_1$ flux. These UAVs must be positioned so that the centre point of their triangulation is equidistant from every UAV in the formation. This is to ensure the final formation position is centred over the target.}

FG is designed to approach and surround a target positioned outside the hemisphere. The minimum distance to the target occurs in the limit where $S_1$ intersects the target and the flux goes to zero. We define this topology as maximum coverage since every UAV is equidistant from the target. The position of $S_2$ with respect to the target can be fine-tuned by a greedy algorithm if required. It can also be obtained by increasing the flux through $S_2$; however, this approach drastically increases the dimensionality of the planning problem.

\section{Constraints for Formation Planning}
\label{sec:constrain}
In this section, we first discuss the improvements we made to LS to make it more effective for UAV path prediction. Next, we present the new FG method, which also uses flux to generate UAV trajectories.

\subsection{LS Modification}
The drawback of LS is that the formation increases in size at each iteration. This is because a larger surface area also increases the flux. This effect was not seen in previous work since the distance from the deformable cloak to the target was small. To solve this problem, we propose adding a term to Equation \ref{equation:ls-cost} to control the size of the formation:
\begin{align}
    L(\vec{\Delta x}) = \vec{\Delta x}^T\vec{\Delta x} + \alpha(\phi_r - \vec{J}_\phi\vec{\Delta x})^2 +
    \beta(\vec{M_1}\vec{\Delta x})^2 ,
    \label{equation:ls-cost-side}
\end{align}
where $\beta$ is a soft constraint for the position retention term, and $\vec{M_1}$ is calculated as:
\begin{align}
    \vec{M_1} = \begin{pmatrix}
    \vec{M_2} & 0 & 0\\
    0 & \vec{M_2} & 0\\
    0 & 0 & \vec{M_2}
\end{pmatrix} ,
\end{align}
where $\vec{B}$ is:
\begin{align}
    \vec{M_2} = \begin{pmatrix}
    1 & -1 & 0 & 0\\
    0 & 1 & -1 & 0\\
    0 & 0 & 1 & -1\\
    -1 & 0 & 0 & 1
\end{pmatrix},
\end{align}

The rows of $\vec{M_2}$ define the linear combinations of the positional changes of the UAVs in the forward-facing group, which in our example are $\vec{\Delta{x}}_1$, ..., $\vec{\Delta{x}}_4$, such that the change in the components of the adjacent UAVs in Figure \ref{fig:hemisphere} are equally constrained. For example, the change in the $x$ component of UAV $\chi_1$ should be equal to the change in the $x$ component of UAV $\chi_2$.

Differentiating Equation \ref{equation:ls-cost-side} and setting the result equal to zero produce the following linear problem:
\begin{equation}
    (1 + \alpha \vec{J}_{\phi}^T \vec{J}_{\phi} + \beta \vec{M_1}^T\vec{M_1})\vec{\Delta x} = \alpha \vec{J}_{\phi}^T \phi_r
    \label{eq:linear-problem-flux}
\end{equation}

\subsection{Target Modelling}
FG considers two types of targets. The first is a single target and the second is a distribution of discrete targets. These target types are useful for tracking objects that are not continuously distributed, such as swarms of UAVs. However, for geometrically complex targets, an approximate charge distribution can be determined by programming an equivalent set of discrete charges that produces an equipotential at the surface of the target \cite{Wang2008a}.

A single target is modelled as a single electric charge and therefore, the contribution of the flux through a triangle in $S_1$ is given by Equation \ref{eq:flux-flux-triangle}. For multiple targets, it is possible to use several point charges and sum the contribution of each to the flux through each triangle. However, a more efficient approach is to consider the centre of charge of the targets and determine an effective radius of the formation based on the maximum relative distance of a target in the swarm with respect to the centre of the charge. In the centre of charge case, the objective function is faster to compute since only a single charge contributes to the total flux.

\subsection{The Flux Guided Method}
LS finds the iterative solution to a regularised linear least-squares problem to solve the path planning problem. In contrast, we solve the flux minimisation problem directly without the least-squares cost function assumption in FG. This eliminates the idea of a target flux value and allows the introduction of constraints for the UAV formation positions.

\textcolor{black}{The goal of FG is to find reference paths for each of the leaders to follow. These paths should not collide and should maximise the flux passing through a surface they triangulate. The first step to calculating the trajectories is to obtain the flux through $S1$. The next step is to find the new positions of the leaders that minimise a local approximation of the flux $\Phi_2({x})$. We define the problem as obtaining the minimum of the negative flux. Specifically, for a surface normal $\hat{n}$ as in Figure \ref{fig:hemisphere}, the flux is negative and decreases in the negative direction as the surface moves toward the target. When the unit vector is reversed, the flux is positive but has the same magnitude. Therefore, by finding the negative flux, we obtain a solution in which the formation is orientated toward the target. At each iteration, a solution to the local approximation of the field is guaranteed provided that a suitable trust-region is constructed. This is because the flux field is differentiable with respect to the positions of the leaders. The only exception is the case where the target lies in the plane of the leaders. We discuss this problem in Section \ref{sec:ReduceDim}.}

Once an expression of the total flux through the surface $S_1$ is found, the positions of the UAVs that increase the flux through the surface can be found by minimising the function $\Phi_2({x})$. In particular, we define the problem as obtaining the minimum of the negative flux. Specifically, for a surface normal $\hat{n}$ as in Figure \ref{fig:hemisphere}, the flux is negative and decreases in the negative direction as the surface moves toward the target. When the unit vector is reversed the flux is positive but has the same magnitude. Therefore, by finding the negative flux we obtain a solution in which the formation is orientated toward the target.

Many potential solutions to the flux minimisation problem are not suitable for real-world applications such as hostile UAV threat mitigation. For instance, the flux can be increased by simply expanding the formation. This solution is not suitable since we require the formation to surround the formation at a given distance. %Also, many local solutions about the target exist. For example, any hemisphere rotated about the target is one of the local solutions.

We find a useful solution by applying constraints to limit the solution to a feasible set defined by the geometry of the formation. As our primary objective is to limit the size of the formation, we propose to impose quadratic equality constraints that limit the distance between adjacent UAVs. Although these constraints allow a rhombic formation of the leaders, the overall flux minimisation should ensure that the formation remains square since the flux is greatest when the formation area is maximised. More stringent conditions were placed on the dot products of the vectors between the leader UAVs; however, these were overly restrictive as they resulted in extremely slow convergence.

\textcolor{black}{We formulate the flux minimisation problem as:
\begin{equation}
\begin{aligned}
    & \underset{\vec{x} \in \R^{12}}{\text{minimize}}
    & & \iint_{S_2(\vec{x})} \vec{E} \cdot \hat{n} \,dA\
\end{aligned}
\end{equation}
}
\textcolor{black}{where $\vec{x}$ is the concatenation of the components of $\vec{x}_1$, $\vec{x}_2$, $\vec{x}_3$, $\vec{x}_4$.}

\textcolor{black}{Using our example setup shown in Figure \ref{fig:hemisphere}, the minimisation has the following constraints on UAVs on $S_2$. We use $\vec{x}$ to clarify the distance measures between each adjacent UAV:
\begin{equation}
\begin{aligned}
%& \text{subject to}
&&& (\vec{x}_1 - \vec{x}_2)^T(\vec{x}_1 - \vec{x}_2) = l^2 \\
&&& (\vec{x}_2 - \vec{x}_3)^T(\vec{x}_2 - \vec{x}_3) = l^2 \\
&&& (\vec{x}_3 - \vec{x}_4)^T(\vec{x}_3 - \vec{x}_4) = l^2\\
&&& (\vec{x}_4 - \vec{x}_1)^T(\vec{x}_4 - \vec{x}_1) = l^2 ,
\end{aligned}
\label{equation:minimisation}
\end{equation}
%where
%\begin{equation}
%    \vec{x}_1 = (x_1, x_2, x_3)^T
%\end{equation}
%\begin{equation}
%    \vec{x}_2 = (x_4, x_5, x_6)^T
%\end{equation}
%\begin{equation}
%    \vec{x}_3 = (x_7, x_8, x_9)^T
%\end{equation}
%\begin{equation}
%    \vec{x}_4 = (x_{10}, x_{11}, x_{12})^T,
%\end{equation}
where $\vec{x}_1, \vec{x}_2, \vec{x}_3, \vec{x}_4$ are the position vectors of leader UAVs in Figure \ref{fig:hemisphere}, %and $x_1$,..., $x_{12}$ are the components of $\vec{x}$, 
and $l$ is the proposed distance between adjacent UAVs.}

The primary advantage of this constraint formulation is the absolute length of formation edges is preserved throughout the minimisation. This means the solution at each iteration is collision-free. Moreover, since the relative motion of the formation is not constrained, the system can perform complex manoeuvres to align itself with and navigate toward the target. Also, the final size of the formation can be set by specifying the final side-length, $l$ in the constraints.

\begin{figure}
    \centering
    \includegraphics[width=0.7\columnwidth]{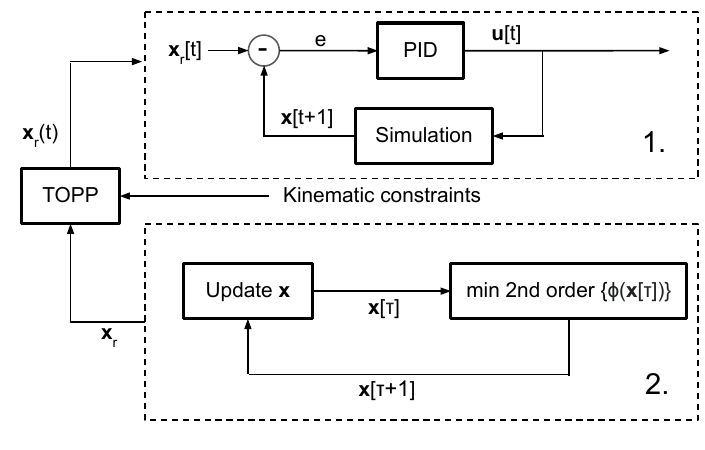}
    \caption{\textcolor{black}{A flow chart describing the main components in the flux guiding method. Component 1 describes the update procedure for generating the trajectories for the leaders. Component 2 describes the controller which uses the trajectories as a reference input to generate a force, $\vec{u}$, to guide a 3d particle simulation. The components are synchronous. First, component 1 generates the trajectories and then component 2 uses the trajectories as the input to the P.I.D controller. $\vec{x}_r$ is the geometric path generated by FG. $\vec{x}_r(t)$ is the trajectory generated by TOPP, $\vec{x}[\tau]$ are the UAV positions at iteration $\tau$ of FG optimisation, $\Phi(\vec{x}[\tau])$ is quadratic approximation of the flux through $S_1$ at step $\tau$, $\vec{x}[t]$ is the simulated position of the UAVs after the application of $\vec{u}[t]$ at time-step $t$, $e$ is the error between the trajectory and the simulated position.}}
    \label{fig:flow-chart}
\end{figure}

\textcolor{black}{Equation \ref{equation:minimisation} is a non-linear optimisation problem with quadratic constraints. Since the objective function is smooth with respect to the positions of the leaders, we can find its minimum by iteratively minimising second-order Taylor approximations about successive leader positions. Since we require the solution to satisfy equality constraints, we can minimise the objective function using Byrd-Omojokun Trust-Region Sequential Quadratic Programming (SQP) \citep{Lalee1998}. %This algorithm is implemented in the SciPy Python package \citep{2020SciPy-NMeth}. 
At each iteration, the trust-region constraints are not necessarily compatible with the linearised equality constraints. Therefore, the algorithm shifts the linear constraint such that a feasible solution can be obtained. In practice, we find that the equality constraints lead to a satisfactory solution for our application. Component 2 in Figure \ref{fig:flow-chart} describes the iterative procedure for calculating $\vec{x}$.}

\section{Experimental Results}
\label{sec:experiment}
We conducted experiments using a computer with an AMD Ryzen 7 3700X CPU and 16GB of RAM. We implemented a software application with Python and standard libraries to simulate the formation of the UAVs under different setups. The computational cost of the algorithm was low and the computational time for the experiments was within a second; a more detailed time cost analysis is included in this section.

The following experiments show the results of the motion planning generated by FG. We also provide additional results in a supplementary video. In each experiment, a hemispherical formation of UAVs is defined by specifying the positions of the four UAVs on the open surface's boundary as shown in Figure \ref{fig:hemisphere}. The target and its radius are specified as inputs to the path planning algorithm. These parameters form the basis of the high-level control mechanism.

Four experiments are set up for different analyses. In Section \ref{sec:exp1}, we compare the paths planned by FG, LS, and LS with an additional soft constraint. In Section \ref{sec:exp2}, we show the results of the formation control problem for tracking a single target. In Section \ref{sec:exp3}, we show efficient multiple target tracking using the centre of charge approach. Finally, in Section \ref{sec:exp4}, we demonstrate the path planning for the entire formation whilst generating paths only for the leader UAVs. 
%The details are presented in the following subsections. 

\subsection{Comparison of Path Planning}
\label{sec:exp1}
In the first experiment, we compare the path planning between the flux guided method and the least-squares approach. A square formation facing in the $\hat{i}$ direction is considered. The initial positions of the UAVs are set to $\vec{x}_1 = (0, 0, 0)$, $\vec{x}_2 = (0, 5, 0)$, $\vec{x}_3 = (0, 5, 5)$, $\vec{x}_4 = (0, 5, 5)$, the target is positioned at $\vec{t} = (40, 40, 40)$, and $\alpha=1000$ and $\beta=0, 400$.

\begin{figure}[h]
\centering
\begin{subfigure}{0.5\linewidth}
\centering
    \includegraphics[width=\linewidth]{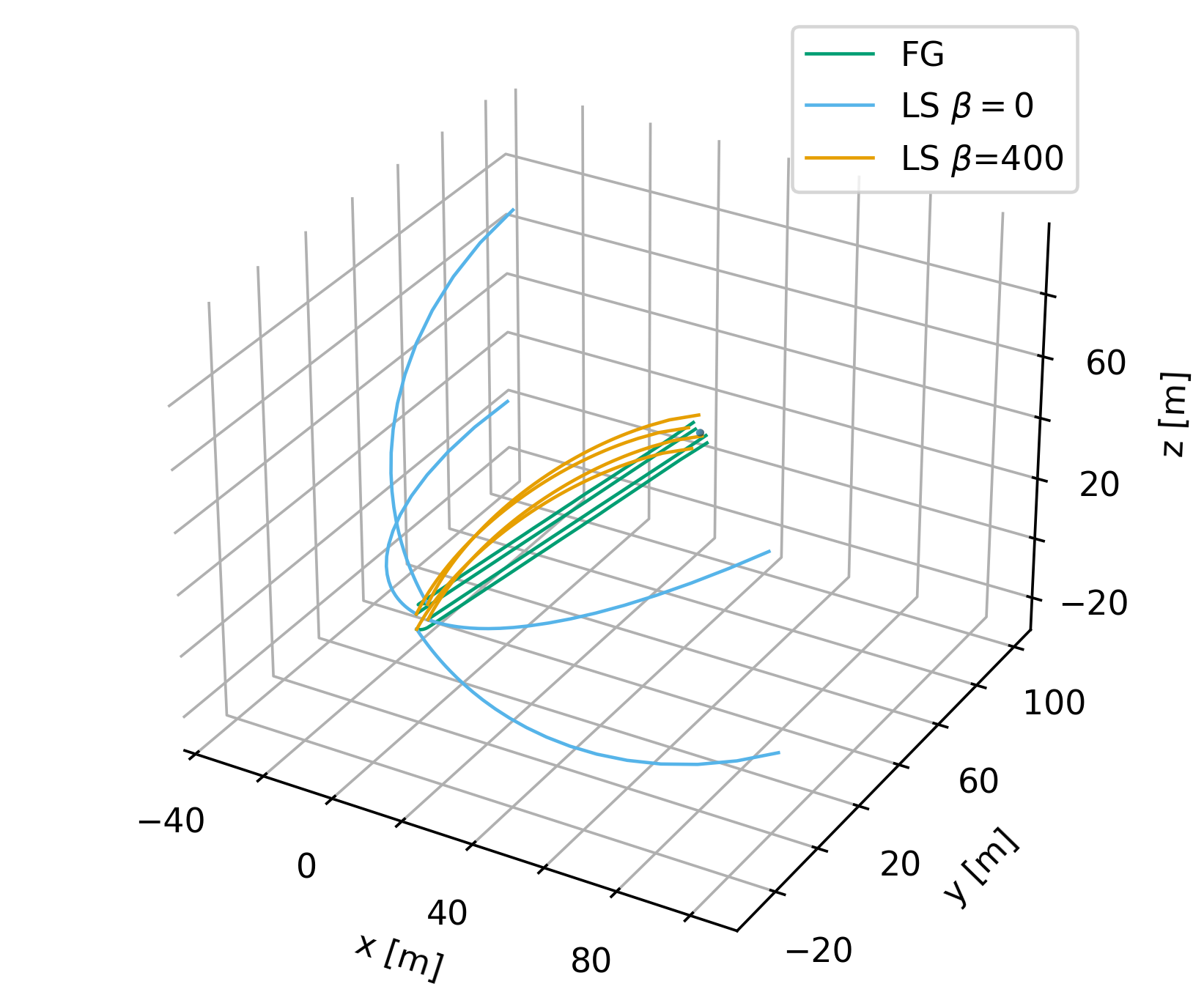}
    \caption{$\vec{t} = (40,40,40)$}%Comparison of path planning for the Flux Guided, and least-squares methods for $\beta=0,400$}
    \label{fig:fg-ls-comparison-1a}
\end{subfigure}%
\begin{subfigure}{0.5\linewidth}
    \centering
    \includegraphics[width=\linewidth]{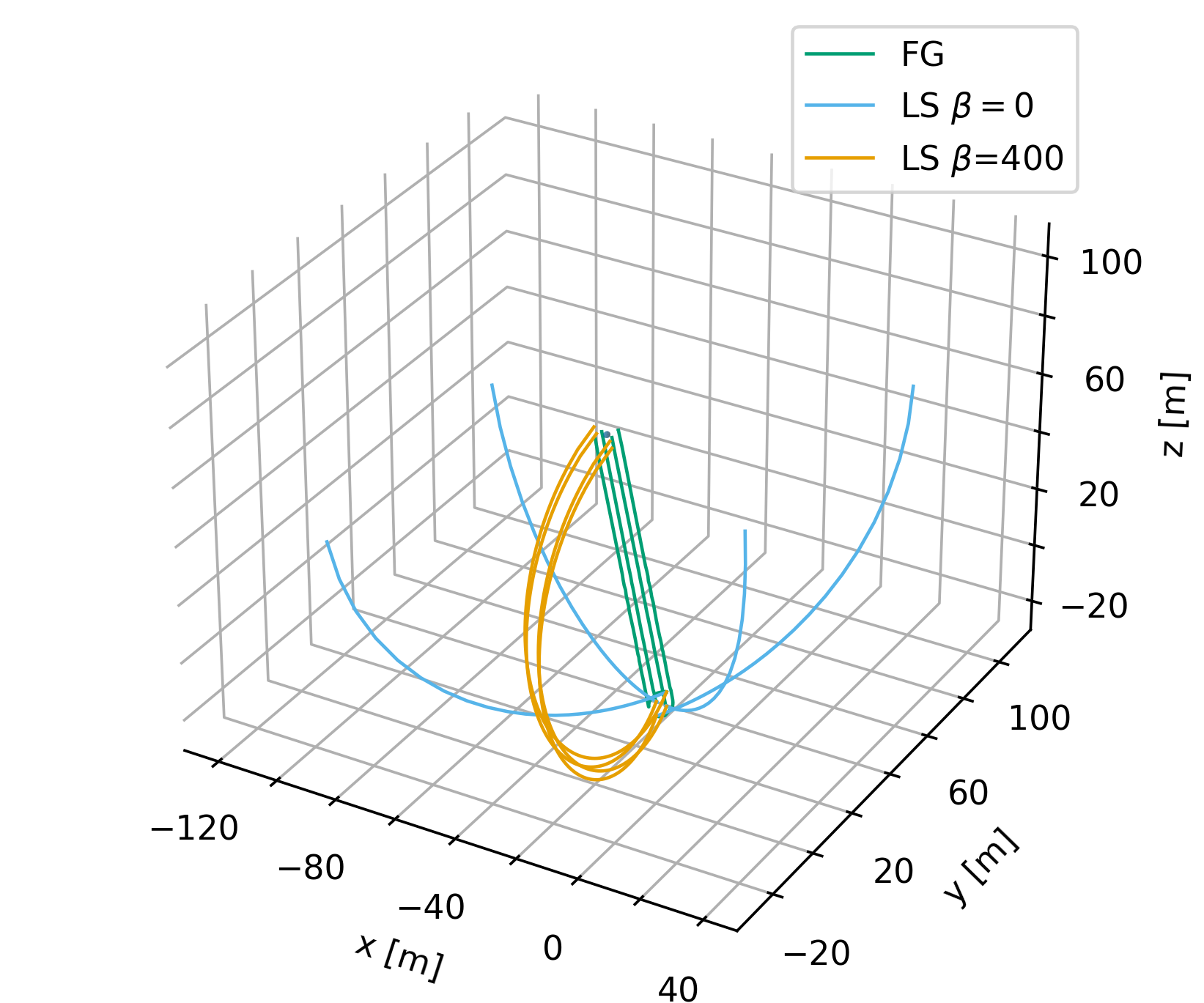}
    \caption{$\vec{t} = (-40,0,0)$}%Comparison of path planning for the Flux Guided, and least-squares methods for $\beta=0,400$}
    \label{fig:fg-ls-comparison-1b}
\end{subfigure}
\caption{Comparison of path planning for the Flux Guided, and least-squares methods for $\beta=0,400$}
\end{figure}

Figure \ref{fig:fg-ls-comparison-1a} shows the paths generated by FG and LS. Each method results in smooth, non-colliding paths which surround the target. Surrounding is defined as the flux through the surface reaching a maximum. LS, $\beta=0$ case yields a path that greatly increases the formation's size. This result is unsuitable where the formation should be maintained in size, or the final positions of the UAVs should be close to the target. Also, for time parameterisation, the traversal time will be greater. In the $\beta=400$ case, the formation retains its shape however the path is arc-shaped. Other values of $\beta$ did not result in straighter paths. The path is arc-shaped because the rotation of the formation is suppressed by the $\beta$ constraint. For example, two opposing UAVs cannot move independently of the two remaining UAVs, since their change in direction must be close to equal. This effect limits the range of motion of the formation significantly. The FG example shows the formation produces an initially rotation and then tracks directly toward the target. This spatially efficient path is possible since only the distance between each UAV is constrained. In addition, straighter paths are preferred since less acceleration is required in comparison with curved paths. Table \ref{table:1} shows the combined path length for each UAV in the formation. LS $\beta=400$ and FG plan paths that are $\approx 100$ \si{\metre} shorter than the original formulation.

In each case, the formation retains its square shape even though the angles between the UAVs are unconstrained. The reason for this is that the square occupies the maximum area for a quadrilateral and therefore the flux through it is a maximum. Where other formation shapes are desired, such as rhombus, the angles between the UAVs must be constrained since the flux through the surface is sub-optimal.

Figure \ref{fig:fg-ls-comparison-1b} shows the planned paths when the target is placed at $\vec{t}=(-40, 0, 0)$. In this case, the formation is facing away from the target. The $\beta=400$ case now performs a more exaggerated arc, and the formation surrounds the target without rotation. In contrast, the formation in FG performs a rotation about $\chi_2$, $\chi_4$ and tracks the target in a straight line. Table \ref{table:1} shows that the FG path is $1.4\times$ shorter than the original formulation, and $1.5\times$ shorter than the soft constrained formulation. This shows that FG is particularly successful when the formation faces away from the target.

\begin{table}[!htb]
\centering
 \begin{tabular}{c| c |c |c} 
 \hline
 Target [\si{\metre}] & LS $\beta=0$ [\si{\metre}] & LS $\beta=400$ [\si{\metre}] & FG [\si{\metre}] \\ [0.5ex] 
 \hline
 (40, 40, 40) & 455 & 346 & 345 \\
 
 (-40, 40, 40) & 500 & 543 & 354  \\ [1ex] 
 \hline
\end{tabular}
\caption{Comparison of the combined path lengths for the UAVs in the square formation given by FG, and LS $\beta=0,400$ path planning algorithms.}
\label{table:1}
\end{table}

\subsection{Tracking a Single Object}
\label{sec:exp2}
In this subsection, we derive time-optimal trajectories with performance constraints for the geometric paths obtained in the previous subsection, subsequently, the trajectories are simulated using a 3d dynamic particle simulation and evaluated against the performance constraints.

FG generates a series of 3d points which form the geometric shape of the paths taken by leader UAVs. Each set of points generated at each iteration of the minimisation is collocated in time. Thus the formation's movement is synchronised. The paths must be re-parameterised in time such that a reference trajectory for a robot can be determined. The re-parametrisation of the path is obtained using a Time-Optimal Path Parameterisation (TOPP) \cite{Pham2018}. TOPP allows performance constraints to be given to the re-parameterisation of the path such that physically realisable trajectories can be obtained. Performance constraints of $|\vec{v}|=10$ \si{\metre\per\second}, and $|\vec{a}|=5$ \si{\metre\per\square\second} are chosen. {These constraints are suitable for quadcopter UAV designs with high manoeuvrability}. Since SQP does not guarantee that the path is smooth some filtering is applied to the path to remove points that are closely spaced prior to the trajectory generation. This was found to improve the efficiency of the TOPP.

\textcolor{black}{An alternative solution to TOPP \cite{Pham2018} is a model predictive trajectory planner \cite{7995821}. This approach finds trajectories directly by maximising the sum of the fluxes over the trajectory with fixed endpoints and kinematic constraints. Whilst this approach obtains trajectories straightforwardly, it greatly increases the number of optimisation variables and is, therefore, a less desirable solution. Also, TOPP is more flexible than MPT since it is decoupled from FG and can be used to generate the follower trajectories.}

To track the trajectory a 3d particle simulation, $\vec{a}=\vec{u}$, is used with the discrete dynamics:
\begin{equation}
    \begin{pmatrix}
    \vec{x}(n+1)\\
    \vec{v}(n+1)\\
    \end{pmatrix}=
    \begin{pmatrix}
    1 & \Delta t\\
    0 & 1\\
    \end{pmatrix}
    \begin{pmatrix}
    \vec{x}(n)\\
    \vec{v}(n)
    \end{pmatrix}+
    \begin{pmatrix}
    \frac{1}{2}\Delta t^2\\
    \Delta t
    \end{pmatrix}\vec{u}(n),
\end{equation}
where $\vec{x}(n)$, $\vec{v}(n)$ are the 3d position and velocity states at $t = n\Delta t$, and $\vec{u}(t)$ is the applied control force. The control force is generated by a PID controller that tracks the trajectory generated by TOPP. Each leader must have access to its global position and the global position of the target to determine and track its trajectory. Alternatively, the UAVs should have access to their positions with respect to the target. However in this case the velocity, and acceleration of the target should be known such that the correct performance constraints can be determined. \textcolor{black}{Figure \ref{fig:flow-chart} shows a flow chart for the components in the flux guiding controller.}

\begin{figure}[h]
\centering
\begin{subfigure}{0.49\linewidth}
\centering
    \includegraphics[width=\linewidth]{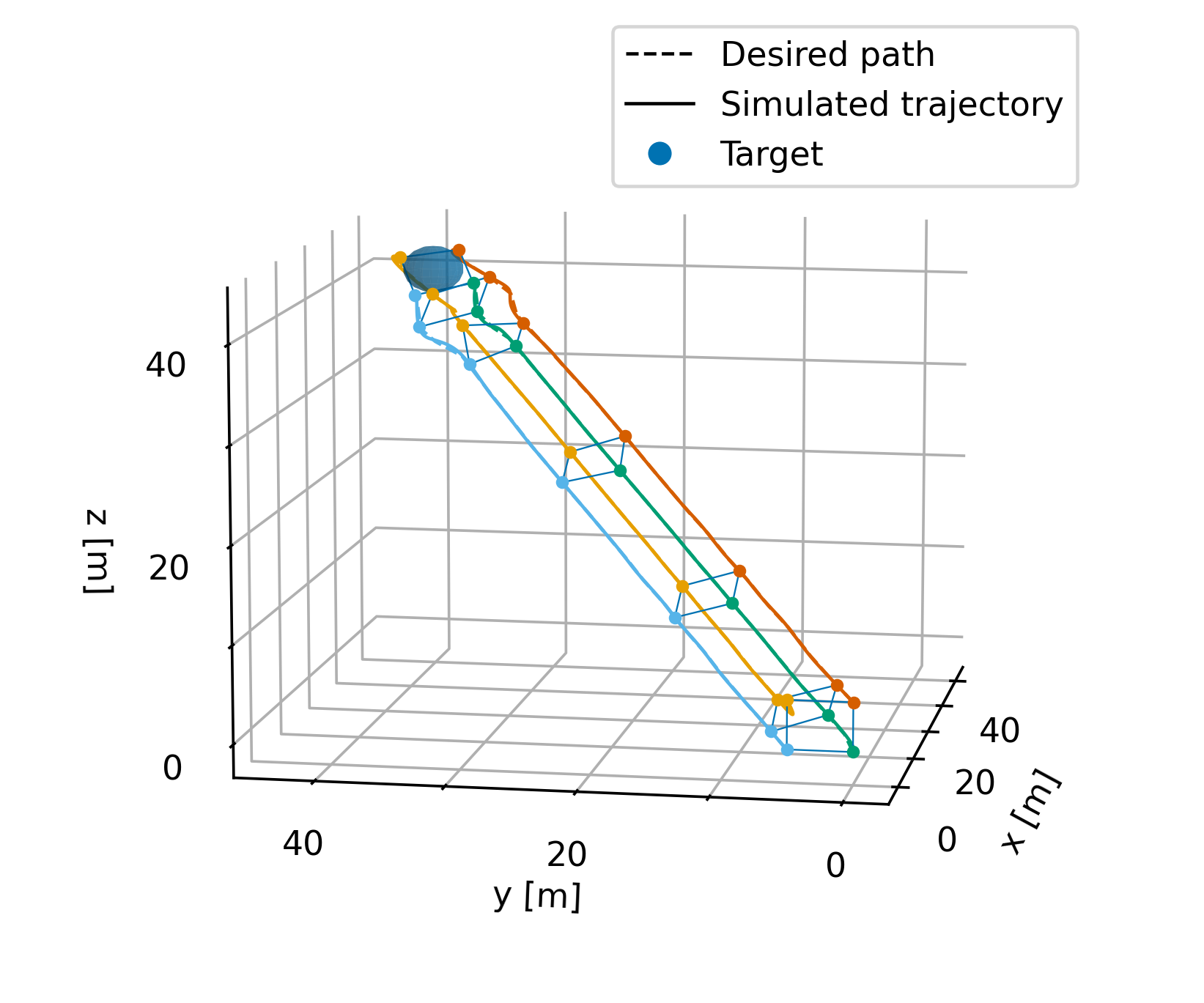}
    \caption{Tracking the target \\$\vec{t} = (40, 40, 40)$}
\end{subfigure}
\centering
\begin{subfigure}{0.49\linewidth} 
\centering
\includegraphics[width=\linewidth]{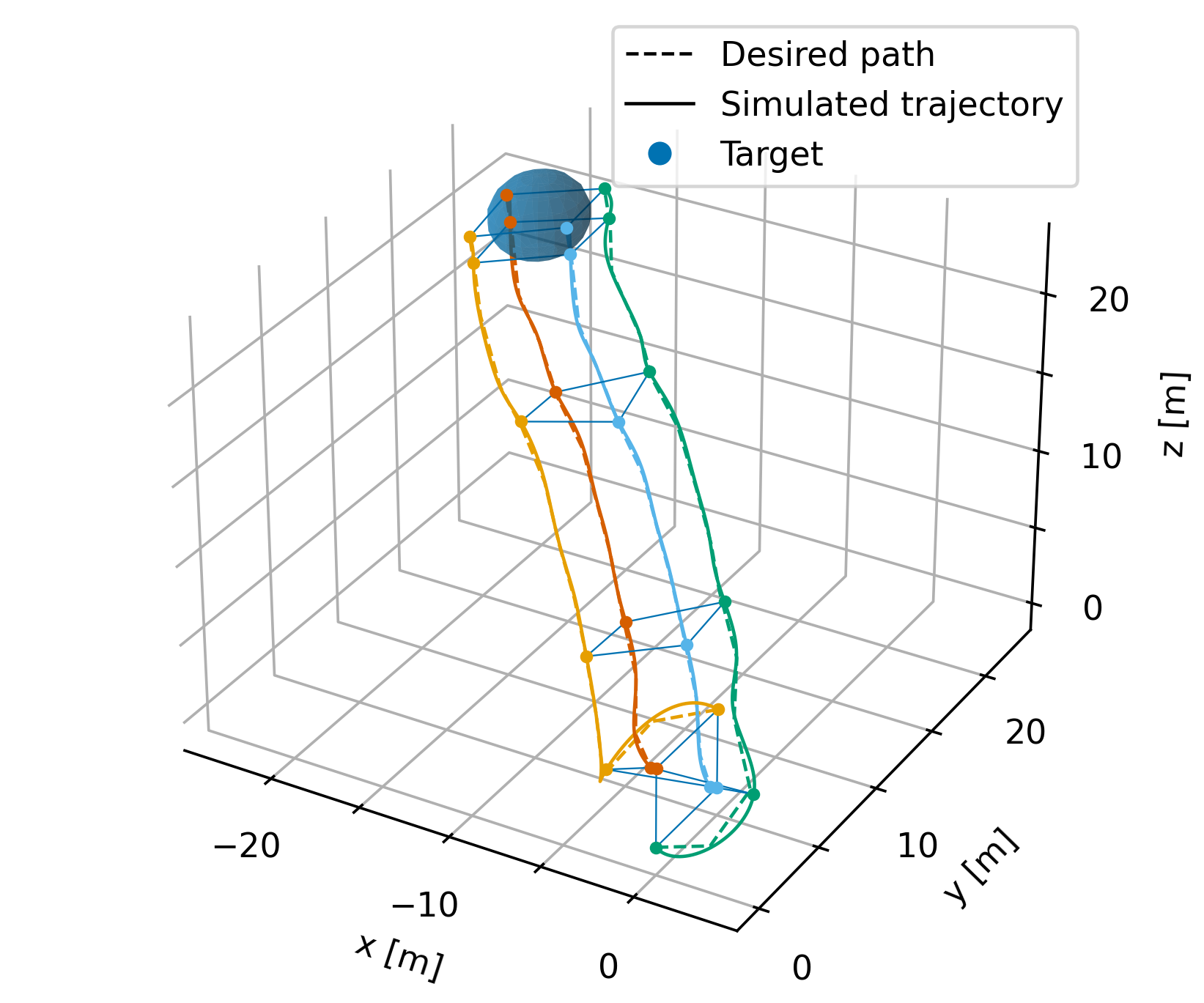}
    \caption{Tracking the target \\$\vec{t} = (-20, 20, 20)$}
\end{subfigure}
 \caption{Comparison of the simulated trajectory and the desired path for leaders in a square formation.}
 \label{fig:2}
\end{figure}

\begin{figure}[h]
    \centering
    \includegraphics[width=0.7\linewidth]{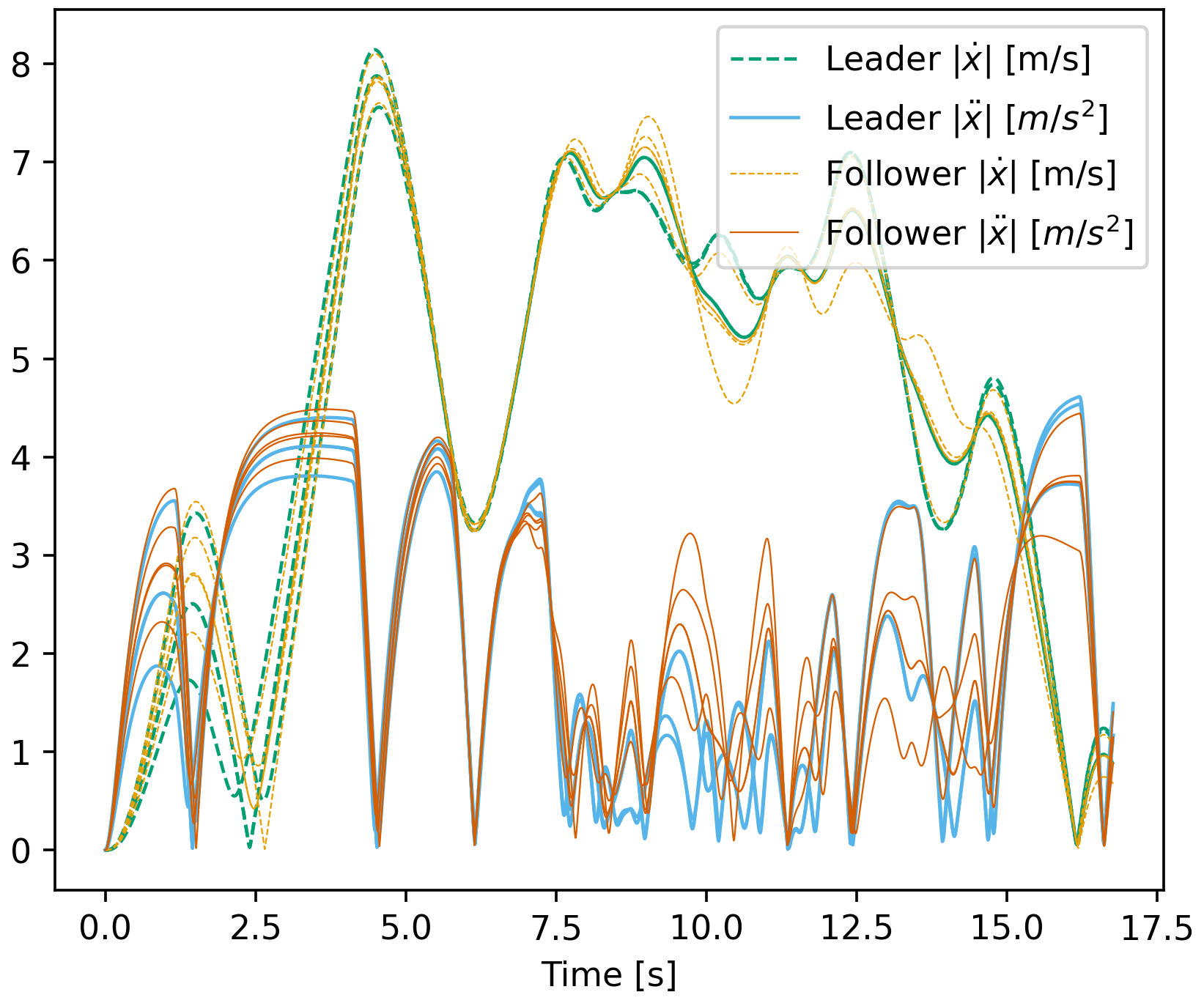}
    \caption{{Simulated velocity and acceleration magnitude curves for four leaders and  5 followers in the formation depicted in Figure \ref{fig:hemisphere} tracking a target positioned at $\vec{t} = (40, 40, 40)$}. The desired trajectories have kinematic constraints $|\vec{v}|=10$ \si{\metre\per\second}, and $|\vec{a}|=5$ \si{m s^{-2}}}.
    \label{fig:2a-v-a}
    
\end{figure}

Figure \ref{fig:2}a shows a comparison of the filtered path generated by FG and the trajectory followed by the particle in the simulation. A square formation is used and the initial positions are the same as in the previous experiments. A target is placed at $\vec{t}=(40, 40, 40)$ \si{\metre}. The figure shows that the particle follows the desired trajectory with little deviation from the desired path. Figure \ref{fig:2a-v-a} shows the velocity and acceleration of each particle throughout the simulation. Since the dynamic model is linear it is expected that the trajectory of the particle will always converge to the desired path if a large enough control force is used. However, the figure shows that the acceleration and hence the control forces are always within the performance constraints. In addition, the velocity curve is within the performance constraints and is relatively smooth. The squares along the trajectory show the shape of the formation during the simulation. The formation remains square throughout the simulation.

\begin{figure}[h]
    \centering
    \includegraphics[width=0.7\linewidth]{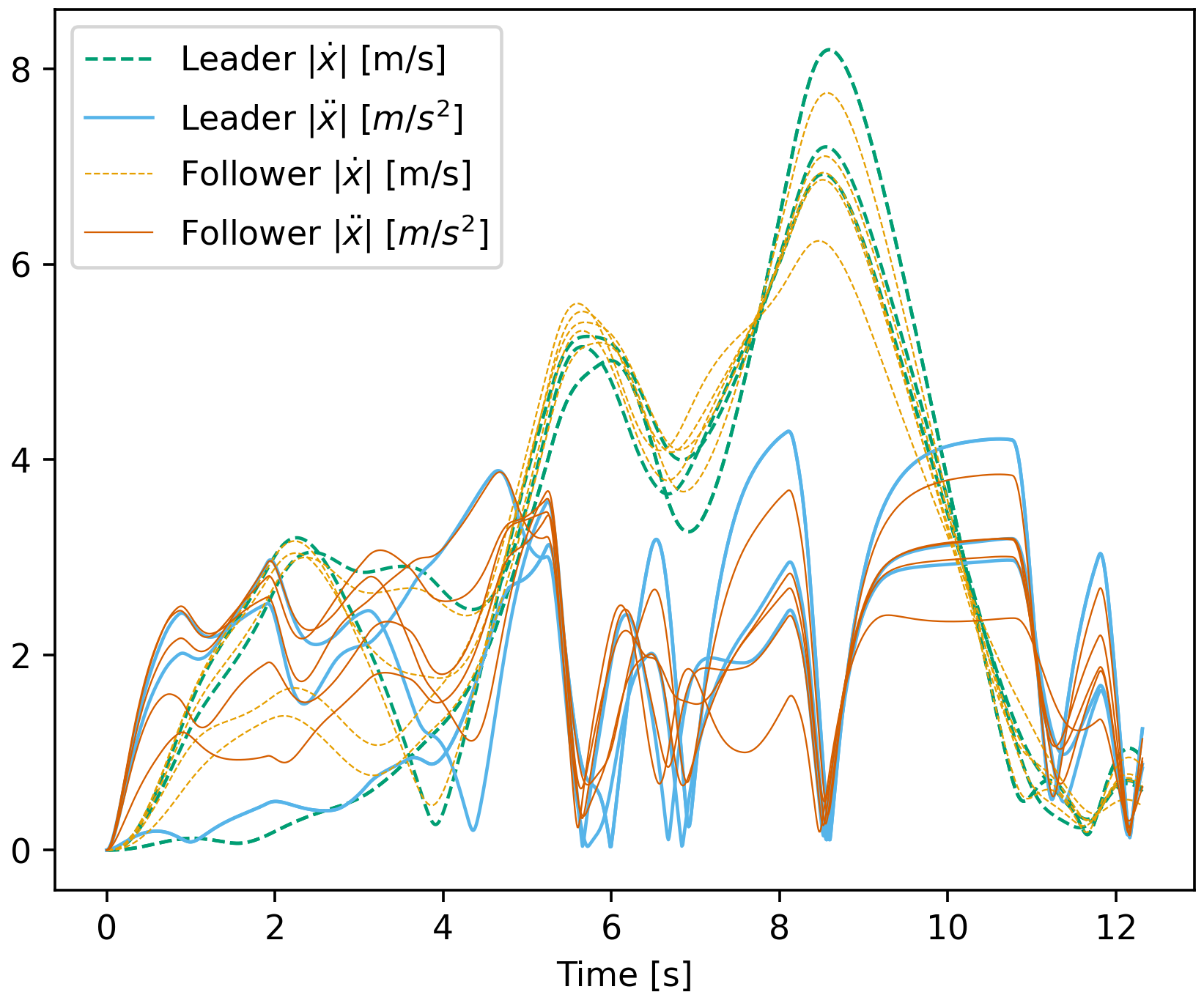}
    \caption{{Simulated velocity and acceleration magnitude curves for four leaders and  5 followers in the formation depicted in Figure \ref{fig:hemisphere} tracking a target positioned at $\vec{t} = (-20, 20, 20)$}. The desired trajectories have kinematic constraints $|\vec{v}|=10$ \si{\metre\per\second}, and $|\vec{a}|=5$ \si{m s^{-2}}}.
    \label{fig:2b-va}
\end{figure}

Figure \ref{fig:2}b repeats the previous experiment with the exception that the target is positioned at $\vec{t}=(-20, 20, 20)$. The figure shows that the trajectories of the particles deviated only slightly from the desired path. The formation remains intact throughout the simulation. In particular, the formation remains square during the rotation manoeuvre at the start of the simulation. Figure \ref{fig:2b-va} shows that the particle remains within the given performance constraints and that the velocity of the particle is quite smooth.

\begin{figure}[h]
\centering
    \includegraphics[width=0.7\linewidth]{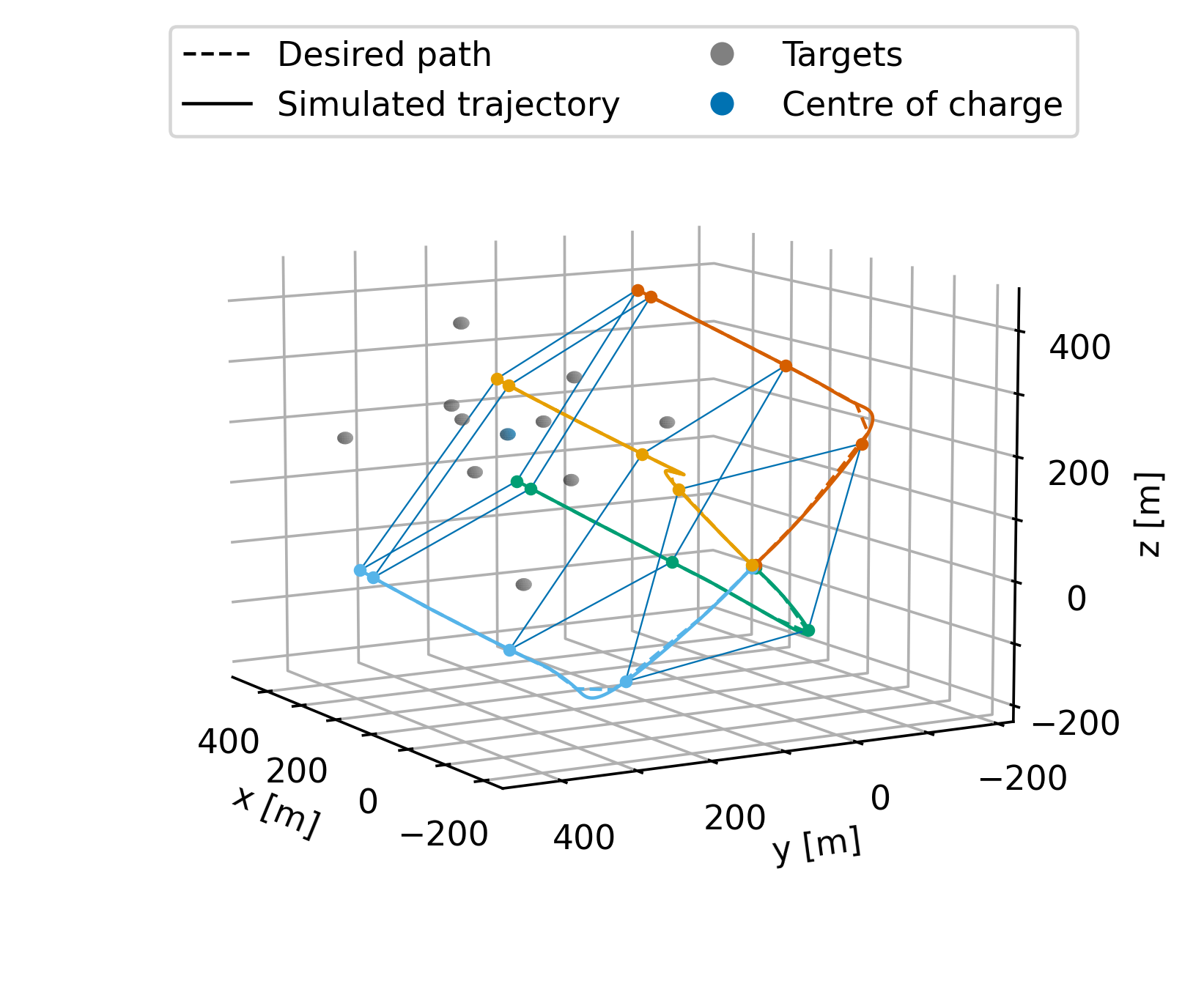}
    \caption{The formation scales and rotates such that targets are surrounded centred at the centre of charge.}
 %\caption{The formation scales and rotates such that targets are surrounded centred at the centre of charge.}
    \label{fig:target-left-up-scaling-multi}
\end{figure}

\subsection{Tracking Multiple Objects}
\label{sec:exp3}
An advantage of FG to LS is that it controls the scale of the formation. An application of this property is that the formation can be guided to cover several targets or guided to the location of a target that has some uncertainty in its position. The approach is computationally efficient since it only considers the location of the centre of charge (COC) as a source of radiation, and an effective radius given by the distance from the outermost target to the COC. The COC is defined as the average of the target positions. To show the flexibility of this approach, the positions of the targets are drawn from a normal distribution. By altering the parameterisation of the distribution we show that FG can respond to varied scenarios or levels of uncertainty in the position of the target.

Figure \ref{fig:target-left-up-scaling-multi} shows the trajectories simulated by FG. The positions of 10 targets are drawn from a normal distribution parameterised by $N(\mu= 200,\sigma= 100)$ shown in grey. The starting positions of the formation are the same as in the previous experiment. The results show that the leaders successfully locate and optimally surround the target formation. The figure shows the formation and the target formation at four time steps. At each time step, the formation retains its shape. The final position of the formation maximises the coverage of the total charge of the targets.

\subsection{Formation Motion}
\label{sec:exp4}
{In comparison to \cite{Ivan2015}, a key advantage of FG is that the degrees of freedom is drastically reduced since only the nodes on the boundary of the open surface area are required for the flux calculation.} This has the result that a complex formation on the boundary can be planned whilst only considering a subset of nodes.

Here, we simulate the motion of the hemisphere shown in Figure \ref{fig:hemisphere} toward a target positioned at $\vec{t} = (10, 10, 10)$. FG plans trajectories for the leader UAVs $\chi_1$, $\chi_2$, $\chi_3$, $\chi_4$. The geometric paths of the follower UAVs are derived from the leaders using vector algebra. In the leader/follower schemes, this is also the case; however, FG can plan trajectories where the formation direction is not in the direction of motion. In most leader/follower schemes, the velocity defines the orientation of the formation. The derivation of these paths is computationally trivial. {Once the paths for the leaders and followers are obtained they are parameterised using TOPP and the previous performance constraints. The TOPP algorithm is decentralised and therefore, it does not increase the computation time of FG.} 

\begin{figure}[!htb]
    \begin{center}
        \includegraphics[width=0.7\linewidth]{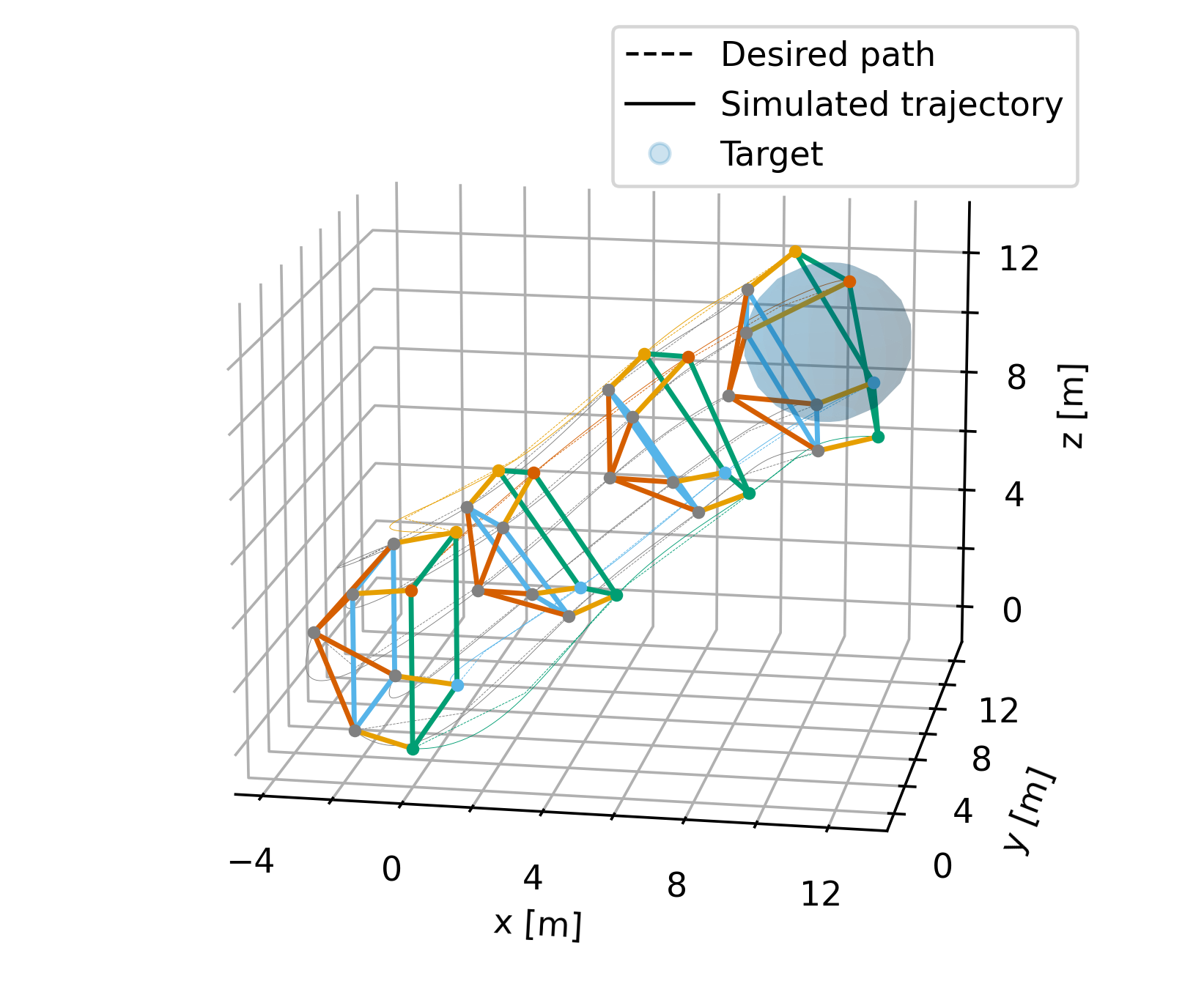}
        \caption{{The simulation of a hemispherical formation surrounding a remote target. The leader UAVs are represented as coloured markers and the followers are in grey. The formation covers the target whilst maintaining its formation. Only the leader UAVs require path planning.}}
        \label{figure:3d1}
    \end{center}
\end{figure}

Figure \ref{figure:3d1} shows the motion of the formation toward the target at the beginning, end and two intermediate stages of the flight. The desired paths and simulated trajectories of each UAV are also shown. The figure shows the formation maintains its shape for the duration of the flight. In addition, the formation optimally surrounds its target. In summary, %addition, tests
the experimental results show that the velocity and acceleration of the leaders are bounded by the performance constraints, and the velocity profile is smooth.

The rotation and translation manoeuvres of the formation appear to be explicitly decoupled. However, FG initially prioritises rotational manoeuvres since they increase the flux more than translation manoeuvres. Very far from the target, $\vec{E}$ is near-uniform, and the change in the dot product between the $\hat{n}$, and $\vec{E}$ is smaller than the change which is obtained by a rotation of $\hat{n}$. Therefore the rotation movement dominates. As the formation approaches the target a mild rotation is coupled with a translation since the field is more non-uniform (Figure \ref{figure:3d1}).

\begin{figure}[!htb]
    \begin{center}
        \includegraphics[width=0.7\linewidth]{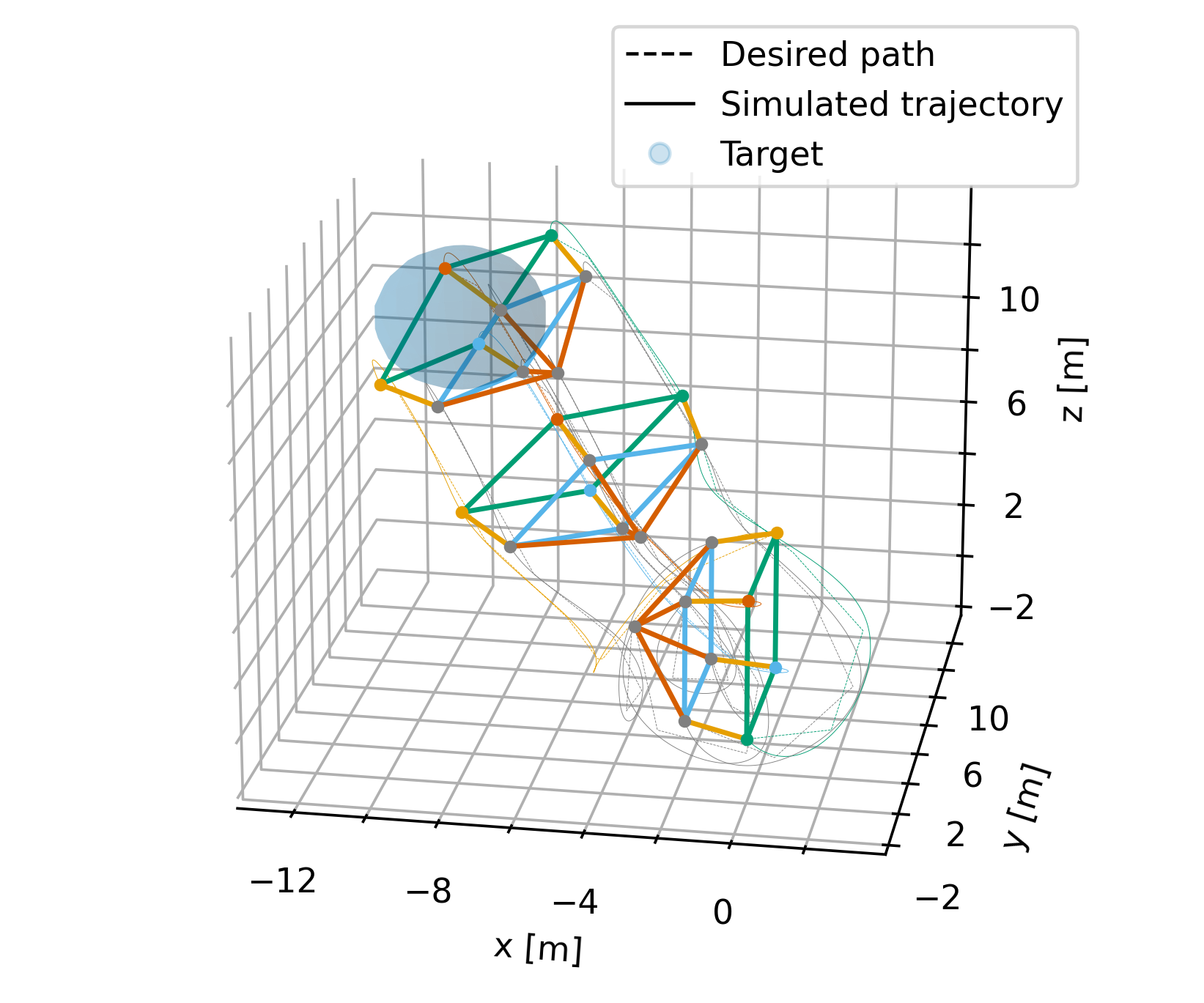}
        \caption{The simulation of a hemispherical formation surrounding a remote target positioned behind the formation's open face. The formation rotates itself and subsequently surrounds the target.}
        \label{figure:3d2}
    \end{center}
\end{figure}

In the next experiment, the same methodology is used and the target is positioned at $\vec{t} = (-10, 10, 10)$. Figure \ref{figure:3d2} shows the formation at three different time steps. The formation retains its shape during the flight and acts to maximise its coverage of the target at all times. Additional tests have shown that the followers perform within the performance constraints and the velocity profile is smooth.

\begin{table}[h!]
\centering
\begin{tabular}{c|c} 
 \hline
 Simulation & Times [\si{\second}]\\ [0.5ex] 
 \hline
 Experiment 1 & 0.52 \\ 
 Experiment 2 & 0.88 \\ [1ex] 
 \hline
\end{tabular}
\caption{The computational time of FG to plan trajectories for a hemisphere of UAVs to track a remote target. Experiment 1 and Experiment 2 are depicted in Figure \ref{figure:3d1}, Figure \ref{figure:3d2} respectively.}
\label{table:execution-times}
\end{table}

Table \ref{table:execution-times} shows the computational times for the experiments leader/follower 1 and leader/follower 2 to plan the geometric paths and perform the TOPP for all leaders and followers in the formation. The execution times are of the order of tenths of a second. The computations have been performed using Python. An algorithm written in a compiled language would likely offer greater performance.

\subsection{\color{black}{FG Sensitivity Test}}
{\color{black}In this experiment, we perform a sensitivity ablation test. As objectives we measure a metric for the flux through the target at the formations final position and the time taken to solve the flux guiding minimisation problem. For the flux metric we choose:
\begin{equation}
    m = \frac{\Phi(\vec{x})}{2\pi}
    \label{eq:flux-metric}
\end{equation}
Equation \ref{eq:flux-metric} is a maximised when half of the flux from the target passing through the formation surface $S_1$. It is allowed that a greater flux passes through the surface, and in this case, we clip the metric at 1. We choose as parameters for the study: the position of the target, and the separation between the UAVs and the target radius. We limit the separation between the UAVs between 5\si{\metre} and 20\si{\metre}, the target between -100\si{\metre} and 100\si{\metre}, and the target radius between 2.5\si{\metre} and 10\si{\metre}. We find these to be reasonable parameters for our use case. We evaluate the objects over 100 random samples drawn from this range. The results are shown in Table \ref{table:sensitivity}.

The average, and standard deviation for the solve time are 1.23 \si{\second}, 1.16 \si{\second} respectively. The average, and standard deviation for the metric are 0.99, 0.03 respectively. These results show that FG consistently finds the optimum positions for the leader UAVs with a reasonable computational time.
% Most likely a compiled version of the optimiser would be faster still.}
}

\begin{table}
\tiny
\centering
\caption{\color{black}{Sensitivity study for the (FG) method. The metric corresponds to the proportion of the optimum flux through the formation's surface at convergence.}}
\label{table:sensitivity}
\begin{tabular}{|c|c|c|c|c|c|c|}
\hline
 Target X [m] &  Target Y [m] &  Target Z [m] &  UAV Sep. [m] &  Target radius [m] &  Metric &  Solve Time [t] \\
\hline
           72 &           -53 &            17 &            11 &                  8 &    0.99 &            0.46 \\
           92 &           -33 &            95 &            16 &                  6 &    1.00 &            1.17 \\
            3 &           -91 &           -79 &            13 &                  3 &    1.00 &            7.21 \\
          -64 &           -13 &           -30 &            16 &                  3 &    1.00 &            1.46 \\
          -12 &            40 &           -42 &             7 &                  3 &    1.00 &            1.03 \\
           93 &           -61 &           -13 &             8 &                  8 &    1.00 &            0.53 \\
           74 &           -12 &           -19 &             5 &                  8 &    0.99 &            0.48 \\
           65 &           -75 &           -23 &             5 &                  9 &    1.00 &            0.32 \\
          -28 &           -91 &            48 &            11 &                  4 &    1.00 &            1.10 \\
           15 &            97 &           -21 &             5 &                  3 &    0.94 &            2.26 \\
           75 &            92 &           -18 &            11 &                  7 &    1.00 &            0.82 \\
           -1 &            77 &           -71 &             8 &                  4 &    1.00 &            2.31 \\
           47 &            47 &            42 &            19 &                  9 &    0.99 &            0.43 \\
           67 &           -68 &            93 &            15 &                  8 &    1.00 &            0.51 \\
          -91 &            85 &            27 &             8 &                  8 &    1.00 &            1.42 \\
          -68 &           -69 &            51 &            13 &                  6 &    0.95 &            1.06 \\
           63 &            14 &            83 &            17 &                  4 &    1.00 &            1.17 \\
          -72 &           -66 &            28 &            13 &                  3 &    0.91 &            2.04 \\
           28 &            64 &           -47 &            18 &                  3 &    0.86 &            1.52 \\
           33 &           -62 &           -83 &            19 &                  7 &    0.99 &            0.79 \\
          -21 &            32 &             5 &            13 &                  6 &    1.00 &            0.52 \\
          -58 &            86 &           -69 &            18 &                  6 &    0.94 &            1.46 \\
           20 &           -99 &           -35 &             7 &                  4 &    1.00 &            2.39 \\
           69 &           -43 &           -65 &             8 &                  9 &    1.00 &            0.40 \\
            2 &            19 &           -89 &            19 &                  8 &    0.99 &            3.23 \\
           74 &           -18 &            -9 &             7 &                  8 &    1.00 &            0.33 \\
           28 &            42 &            -1 &            16 &                  3 &    1.00 &            0.59 \\
          -47 &            40 &            21 &            18 &                  3 &    1.00 &            0.92 \\
           70 &           -16 &           -32 &             5 &                  4 &    1.00 &            0.94 \\
          -94 &            96 &           -53 &            13 &                  5 &    1.00 &            1.38 \\
           27 &            31 &             0 &            13 &                  3 &    0.95 &            0.42 \\
           80 &           -22 &            43 &             8 &                  9 &    0.98 &            0.54 \\
           48 &            86 &           -77 &            13 &                  6 &    1.00 &            1.07 \\
           41 &            17 &           -15 &            15 &                  9 &    1.00 &            0.29 \\
          -52 &           -51 &           -31 &             7 &                  8 &    1.00 &            0.55 \\
           69 &            63 &            92 &            13 &                  4 &    1.00 &            2.29 \\
           -5 &            97 &            -6 &             9 &                  4 &    1.00 &            1.32 \\
         -100 &            13 &            78 &             8 &                  9 &    1.00 &            0.95 \\
          -64 &            62 &           -52 &            17 &                  7 &    0.91 &            0.86 \\
           -7 &            31 &            -2 &            18 &                  3 &    1.00 &            0.36 \\
          -58 &            12 &            49 &            19 &                  8 &    1.00 &            1.01 \\
           27 &          -100 &            38 &             5 &                  3 &    1.00 &            2.32 \\
           14 &           -57 &            86 &             9 &                  6 &    1.00 &            0.93 \\
           27 &           -77 &            87 &             8 &                  5 &    0.96 &            3.95 \\
           30 &            21 &            -2 &            18 &                  6 &    0.99 &            0.31 \\
          -38 &            63 &            23 &            16 &                  5 &    1.00 &            0.61 \\
           95 &           -18 &            74 &            17 &                  7 &    1.00 &            0.76 \\
           48 &           -50 &            55 &            11 &                  9 &    1.00 &            0.48 \\
          -86 &           -59 &           -42 &            18 &                  6 &    0.98 &            1.22 \\
           93 &           -64 &           -90 &            14 &                  9 &    1.00 &            0.89 \\
\hline
\end{tabular}
\end{table}

\section{Conclusions and Discussions}
\label{sec:conclusions}
We have demonstrated a new formulation for the path planning of UAVs using a harmonic electric field. We plan collision-free paths for UAVs in hemispherical formations to encircle targets whilst maintaining the shape of the formation. In particular, our objective function is designed to maximise the coverage of the target throughout the flight. \textcolor{black}{This has the practical advantage that the formation can optimally monitor a target. For example, remote sensors such as vision \cite{issacmedina21unmanned} or LiDAR \cite{li21durlar} maintain a minimally occluded view of their target. Also, projectiles can be fired at remote targets without risk to other UAVs in the formation. This type of coverage is not considered in current leader/follower and encirclement schemes. We have also demonstrated the FG formulation can control the scale of the formation during flight. In a practical setting, this property allows the formation to closely follow its target regardless of its initial scale.} We have demonstrated that smooth trajectories can be generated from the paths. Furthermore, we have shown that the trajectories are suitable for robotic applications by simulating the response of a 3d dynamic particle system under the control of a PID controller. 

The new formulation has several advantages. Firstly, the combined total path length of the UAVs is significantly shorter. Secondly, the scale and shape of the formation can be controlled. This allows groups of targets to be tracked, or uncertainty in the target's location to be taken into account. Also, the scheme retains the novel scalable property that path planning is only required for the UAVs on the open boundary of the formation's surface.

Compared to previous work based on electric flux planning \cite{Wang2013, Ivan2015}, this work proposes a static encirclement scheme only. However, FG can be used to encircle targets in motion. The planning algorithm is updated in an inner loop to the target sensing loop for the target. Operating in this mode motion tracking of the target would not be comprised by the current performance (see Table \ref{table:execution-times}). This is because subsequent iterations of FG converge quickly since the new solution is close to the previous solution.

We showcase that our flux-guided controller is compatible with a leader-follower approach. While the implementation of a decentralised flux-guided controller is out of the scope of this research, it is an interesting future direction. We believe that the formation indicated in this paper can be less formally defined without explicitly classifying UAVs into a forward-facing group and the volume-maintaining group. In particular, in a decentralised setup, each UAV would maintain the relative distance with its surrounding neighbours, thereby forming a dynamic mesh structure similar to the formation controller of \cite{henry14interactive}. UAVs with detectable targets would approach the targets by following the flux as explained in this work, and such movement would pull the formation towards the target, showcasing formation behaviours similar to that of \cite{henry12environment}.

A current limitation of FG is its centralised operation. In future work, the algorithm could be adapted to a decentralised setting. One proposal to mitigate this is to have each leader compute its trajectory using the virtual position of the other leaders defined by the topology of the formation and initial orientation. Similarly, the followers can compute the leader's trajectory and then plan their relative trajectory. We have not included this variation in this work since noise in the global positioning system should be taken into careful consideration to ensure stability in the control algorithm.

%\textcolor{black}{Although a real-world demonstration of formation UAV control is the ultimate goal of this research, it is also dependent on other factors that are not directly related to the control we investigate, such as various hardware, weather condition, etc. It is commonly accepted that control is first studied in simulation environments before being applied to real-world drones for example \cite{Ma2018}. For this paper, our focus is on the new control algorithm. Finally, the trajectory generation system proposed in this work has been developed for the DJI drones API. As such the scheme can be experimentally validated by outdoor testing. The team is currently working on this future direction.}
\textcolor{black}{Although a real-world demonstration of formation UAV control is the ultimate goal of this research, the success would depend on other components that are not directly related to the control we investigate. In particular, we do not touch on problems such as detecting and tracking hostile UAVs, which would be needed for a real-world demonstration. That said, such problems are well studied and existing solutions are available via visual sensors \cite{issacmedina21unmanned,organisciak22uavreid}. This research serves as the first study in simulation environments to showcase the flexibility of the proposed system, focusing on the newly proposed control algorithm. Combining the control system with other components for a real-world demonstration is our ongoing direction.}

\section*{Acknowledgement}
This work is supported by the MOD Chief Scientific Adviser’s Research Programme, through the Defence and Security Accelerator (Ref: DSTLX-1000140725), and the Royal Society (Ref: IES$\backslash$R2$\backslash$181024 and IES$\backslash$R1$\backslash$191147).

\bibliography{library}

\end{document}